\documentclass{article}

\usepackage[main, final]{neurips_2026}
\usepackage{amsmath}
\usepackage{graphicx}
\usepackage{subcaption}
\usepackage{siunitx}
\usepackage{algorithm}
\usepackage{indentfirst}
\usepackage{algpseudocode}
\usepackage{listings}
\usepackage{colortbl}

\usepackage[utf8]{inputenc} 
\usepackage[T1]{fontenc}    
\usepackage{hyperref}       
\usepackage{url}            
\usepackage{booktabs}       
\usepackage{amsfonts}       
\usepackage{nicefrac}       
\usepackage{microtype}      
\usepackage{booktabs}
\usepackage{xcolor}
\usepackage{colortbl}
\usepackage{multirow}         

\title{MCSA: Multi-Cutoff Spectral Aggregation in SO(3) Equivariant GNNs for Interatomic Potentials}

\author{%
  Junfu Tan\textsuperscript{1,2,$\dagger$}, Mengfan Wu\textsuperscript{1,3,$\dagger$}, Yu Zhu\textsuperscript{1,4}, Yu Shi\textsuperscript{5,6}, Yifang Qin\textsuperscript{7},  \\
  \textbf{Chang Liu\textsuperscript{5,6,$*$}, Bonan Zhu\textsuperscript{8,9,$*$}, Ziheng Lu\textsuperscript{1,5,6,$*$}}  \\
  \textsuperscript{1}{Kairos Materials} 
  \textsuperscript{2}{The University of Hong Kong} 
  \textsuperscript{3}{Tongji University} \\ 
  \textsuperscript{4}{Westlake University} 
  \textsuperscript{5}{Zhongguancun Institute of Artificial Intelligence}\\ 
  \textsuperscript{6}{Zhongguancun Academy}
  \textsuperscript{7}{Peking University }
  \textsuperscript{8}{Beijing Institute of Technology}\\ 
  \textsuperscript{9}{State Key Laboratory of Environment Characteristics and Effects for Near-space}
  \\ \textsuperscript{$\dagger$}{Equal contribution.} \textsuperscript{$*$}{Corresponding author.} \\ 
  \texttt{junfu.tan@connect.hku.hk},
  \texttt{liuchang@bza.edu.cn},\\
  \texttt{bzhu@bit.edu.cn},
  \texttt{zluag@connect.ust.hk}
}
\begin{document}

\maketitle

\begin{abstract}
  Machine Learning Interatomic Potentials (MLIPs) based on Equivariant Graph Neural Networks (EGNNs) have achieved remarkable progress in atomistic modeling. However, accurately representing atomic interactions across different spatial scales remains challenging. In this work, we investigate the radial spectral properties of SO(3)-equivariant message passing and establish that the cutoff radius controls the characteristic spatial-frequency scale of radial interactions. Based on this analysis, we propose Multi-Cutoff Spectral Aggregation (MCSA), a plug-and-play module that aggregates radial kernels with different cutoff radii to enrich multi-scale geometric representations. To achieve efficient implementation, we exploit the compact support of radial kernels to fuse cutoff-dependent radial weights before angular tensor-product operations, avoiding redundant angular computations while preserving rotational equivariance. Extensive experiments across representative EGNN backbones demonstrate consistent improvements in energy, force, and stress prediction, as well as enhanced accuracy in higher-order properties, including phonon dispersion and lattice thermal conductivity. Further evaluations on layered materials with distinct interaction scales demonstrate the effectiveness of physically motivated cutoff selection. Code is available at: {\urlstyle{same}\hypersetup{allcolors=magenta}\url{https://github.com/JF-Tan/MCSA}}.
  
\end{abstract}


\section{Introduction}
The quantitative simulation of atomic systems is fundamental to advancements in materials science and drug discovery\cite{bergstrom2018computational, borhani2012future, karplus1990molecular, deng2015quantitative}, yet it has long been constrained by a dichotomy between the accuracy of high-complexity ab initio methods\cite{dftKohnSham, geerlings2003conceptual, sholl2022density} and the efficiency of limited classical potentials\cite{eam, LJpotential, muser2023interatomic, balamane1992comparative, hata2017development}. Recently, Machine Learning Interatomic Potentials (MLIPs) have emerged as a pivotal solution\cite{Batatia2022mace,equiformer_v2,Batzner2022, Musaelian2023Allegro}, effectively bridging this gap by approximating the potential energy surface with neural networks. In particular, Equivariant Graph Neural Networks (EGNNs) \cite{Batatia2022mace,Musaelian2023Allegro,Simeon2022TensorNet,Geiger2022e3nn} have driven a major paradigm shift in atomistic modeling. By preserving physical and geometric symmetries, EGNNs achieve remarkable data efficiency and force prediction fidelity, enabling reliable molecular dynamics simulations across complex, heterogeneous systems.

Despite these remarkable advancements, the fine-grained representation of structural geometry remains an ongoing challenge in the development of robust MLIPs. Significant progress has been made in the angular domain, where the introduction of higher-order spherical harmonics and Clebsch-Gordan tensor products has yielded highly refined geometric representations \cite{Geiger2022e3nn, Batzner2022}. Concurrently, in the radial domain, the adoption of radial basis functions (e.g., Gaussian or Bessel functions) has further improved spatial resolution \cite{esen, behler2007generalized, schutt2018schnet, gasteiger2020dimenet}. However, a fundamental limitation in current EGNN architectures is frequently overlooked: the reliance on a single cutoff radius for all message-passing interactions.

In this work, we analyze the radial spectral response of SO(3)-equivariant message-passing networks and show that the cutoff radius governs the characteristic spatial-frequency scale of radial interactions. Under a scale-consistent radial parameterization, a single cutoff provides a single characteristic spectral scale, potentially limiting the representation of fine-grained local geometries alongside broader atomic interactions. To address this limitation, we propose \textbf{Multi-Cutoff Spectral Aggregation (MCSA)}, a plug-and-play module that combines radial kernels with different cutoff radii. By superposing their differently scaled frequency responses, MCSA enriches the radial representation across multiple spatial scales while preserving the underlying equivariant structure.

Furthermore, naively implementing multiple cutoffs may incur redundant evaluations of computationally expensive angular tensor products. We address this challenge through an efficient radial-fusion formulation that exploits the compact support of cutoff radial kernels to unify neighborhood aggregation. Specifically, the cutoff-dependent radial weights are combined before the angular tensor-product operation, allowing each edge and coupling path to share a single angular computation across all cutoffs. This formulation avoids redundant angular tensor products while retaining the original multi-cutoff aggregation, with additional computational cost arising primarily from radial-weight evaluation and fusion.

To comprehensively evaluate MCSA, we integrate it into multiple representative SO(3)-equivariant GNN backbones. Experiments demonstrate consistent improvements in energy, force, and stress prediction, together with enhanced accuracy in higher-order physical properties, including phonon dispersion and lattice thermal conductivity. We further investigate layered materials, where short-range in-plane covalent bonds coexist with longer-range interlayer interactions, providing a targeted evaluation of MCSA in structurally heterogeneous environments. Finally, we evaluate PES smoothness and long-term molecular dynamics stability, demonstrating that multi-cutoff aggregation maintains smooth energy and force profiles without introducing artificial cutoff-boundary artifacts.

\section{Preliminary: SO(3) Equivariant Message Passing}
Modern Machine Learning Interatomic Potentials (MLIPs) typically employ SO(3) equivariant message-passing mechanisms \cite{Batatia2022mace,equiformer_v2,Batzner2022}, where node features are composed of \textit{irreducible representations} (Irreps) of the SO(3) group, allowing geometric information to be propagated while preserving rotational equivariance.
Consider a system of $N_{\mathrm{at}}$ atoms with coordinates
$\{\mathbf R_i\}_{i=1}^{N_{\mathrm{at}}}$.
Let
$
\mathbf r_{ij}=\mathbf R_j-\mathbf R_i,
$
$
\hat{\mathbf r}_{ij}=\frac{\mathbf r_{ij}}{\|\mathbf r_{ij}\|}$,
denote the relative position and corresponding unit direction, respectively. Define $\mathcal N(i)=\{j:\|\mathbf r_{ij}\|<R_c\}$ as the neighborhood of atom $i$ within a cutoff radius $R_c$. A general SO(3) equivariant message-passing layer can be expressed as
\begin{equation}
\label{eq:so3_mp}
\mathbf x_i'^{(l_3)}
=
\sum_{j\in\mathcal N(i)}
\sum_{l_1,l_2}
\mathbf x_j^{(l_1)}
\otimes_{l_1,l_2}^{l_3}
Y^{(l_2)}(\hat{\mathbf r}_{ij})
\Phi_{R_c}^{(l_1,l_2,l_3)}(\|\mathbf r_{ij}\|),
\end{equation}
where $\mathbf x_j^{(l_1)}\in\mathbb R^{2l_1+1}$ denotes the
degree-$l_1$ feature of atom $j$, and
$Y^{(l_2)}(\hat{\mathbf r}_{ij})$ collects the spherical
harmonics of degree $l_2$. We adopt real orthonormal spherical harmonics throughout
the derivation. The radial function
$\Phi_{R_c}^{(l_1,l_2,l_3)}(\|\mathbf r_{ij}\|)$
determines the distance-dependent interaction strength for
each angular coupling pathway and is parameterized with respect
to the cutoff radius $R_c$. It is typically the product of a radial basis function and a smooth truncation function.
The operator $\otimes_{l_1,l_2}^{l_3}$ denotes the
Clebsch--Gordan (CG) tensor product, which couples the
irreducible representations of degrees $l_1$ and $l_2$
and projects their tensor product onto the output irrep $l_3$,
subject to the angular momentum selection rules
$|l_1-l_2|\leq l_3\leq l_1+l_2$.
The radial envelope is designed to ensure that the
interaction smoothly vanishes at the cutoff boundary,
with sufficient differentiability to support the
evaluation of energy and force derivatives \cite{esen}. Expanding the CG projection in components gives
\begin{equation}
\label{eq:so3_mp_components}
\begin{aligned}
\mathbf x_i'^{(l_3,m_3)}
={}&
\sum_{l_1,l_2}
\sum_{m_1,m_2}
C_{l_1m_1,l_2m_2}^{l_3m_3}
\\
&\quad\cdot
\sum_{j\in\mathcal N(i)}
\mathbf  x_j^{(l_1,m_1)}
 Y^{(l_2,m_2)}(\hat{\mathbf r}_{ij})
\Phi_{R_c}^{(l_1,l_2,l_3)}(\|\mathbf r_{ij}\|),
\end{aligned}
\end{equation}
where $C_{l_1m_1,l_2m_2}^{l_3m_3}$ denotes a CG coefficient. Eq.~\eqref{eq:so3_mp_components} separates the angular and
radial contributions to message passing. The angular dependence
is carried by spherical harmonics, whereas the distance-dependent
interaction strength is controlled by the radial weight
$\Phi_{R_c}^{(l_1,l_2,l_3)}$.
This angular-radial separation provides a natural basis
for analyzing the spectral properties of the radial kernel,
particularly how the cutoff radius $R_c$ shapes its
represented spatial frequency components.

\section{Radial Spectral Response in SO(3)-Equivariant Message Passing}

To characterize the spatial-frequency response of the radial
kernel, we first reformulate the discrete message aggregation
as a continuous spatial convolution. Specifically, we define
the global atomic feature distribution as
\begin{equation}
\rho^{(l_1,m_1)}(\mathbf y)
=
\sum_{j=1}^{N_{\mathrm{at}}}
\mathbf x_j^{(l_1,m_1)}
\delta(\mathbf y-\mathbf R_j),
\end{equation}
where $\delta(\cdot)$ denotes the Dirac delta distribution.
We further define the spatial aggregation kernel as
\begin{equation}
K_{R_c}^{(l_1,l_2,l_3,m_2)}(\mathbf r)
=
\Phi_{R_c}^{(l_1,l_2,l_3)}(\|\mathbf r\|)
Y^{(l_2,m_2)}(\hat{\mathbf r}),
\end{equation}
with compact support
within the cutoff radius $R_c$.
The corresponding continuous feature aggregation is then
expressed as
\begin{equation}
\begin{aligned}
x'^{(l_3,m_3)}(\mathbf p)
={}&
\sum_{l_1,l_2}
\sum_{m_1,m_2}
C_{l_1m_1,l_2m_2}^{l_3m_3}
\\
&\cdot
\int_{\mathbb R^3}
\rho^{(l_1,m_1)}(\mathbf y)
K_{R_c}^{(l_1,l_2,l_3,m_2)}
(\mathbf y-\mathbf p)
\,d\mathbf y,
\end{aligned}
\end{equation}
where $\mathbf p$ denotes an arbitrary spatial position,
and the original node-wise aggregation is recovered by
evaluating the continuous feature field at
$\mathbf p=\mathbf R_i$.
This formulation separates the atomic feature distribution
from the spatial aggregation kernel, providing a convenient
basis for analyzing the spectral response of the
message-passing operation.

Define the spatial kernel
$K_{R_c}(\mathbf r)=\Phi_{R_c}^{(l_1,l_2,l_3)}(\|\mathbf r\|)Y_{l_2m_2}(\hat{\mathbf r})$.
Under the Fourier transform convention
$\widehat f(\mathbf k)=\int_{\mathbb R^3}f(\mathbf r)e^{-i\mathbf k\cdot\mathbf r}\,d\mathbf r$,
the plane-wave expansion \cite{jackson1998classical} gives
\begin{equation}
\widehat K_{R_c}(\mathbf k)
=
(-i)^{l_2}Y_{l_2m_2}(\hat{\mathbf k})
H_{R_c}^{(l_1,l_2,l_3)}(q),
\end{equation}
where $q=\|\mathbf k\|$ and
\begin{equation}
\label{eq:radial_response}
H_{R_c}^{(l_1,l_2,l_3)}(q)
=
4\pi\int_0^{R_c}
r^2\Phi_{R_c}^{(l_1,l_2,l_3)}(r)
j_{l_2}(q r)\,dr.
\end{equation}
Here, $j_{l_2}$ is the spherical Bessel function of the first kind. 
Since the aggregation uses $K_{R_c}(\mathbf y-\mathbf p)$,
the corresponding Fourier multiplier is evaluated at
$-\mathbf k$ rather than $\mathbf k$. This sign reversal
introduces only an angular phase change and does not
affect the radial response $H_{R_c}^{(l_1,l_2,l_3)}(q)$.

We now assume a scale-consistent radial parameterization
\begin{equation}
\Phi_{R_c}^{(l_1,l_2,l_3)}(r)
=
a_{R_c}^{(l_1,l_2,l_3)}
\psi^{(l_1,l_2,l_3)}(r/R_c),
\end{equation}
where  $a_{R_c}^{(l_1,l_2,l_3)}$ is a cutoff-dependent amplitude, and $\psi^{(l_1,l_2,l_3)}$ is a dimensionless radial profile defined on the normalized interval $[0,1]$.
In particular, the dependence on $r/R_c$ ensures that the normalized radial profile remains unchanged as the cutoff radius varies, while its spatial support scales with $R_c$. This parameterization therefore isolates the effect of the cutoff radius on the spatial scale of the radial kernel. Substituting $s=r/R_c$ into Eq.~\eqref{eq:radial_response} gives
\begin{equation}
\label{eq:cutoff_scaling}
H_{R_c}^{(l_1,l_2,l_3)}(q)
=
a_{R_c}^{(l_1,l_2,l_3)}R_c^3
H^{(l_1,l_2,l_3)}(q R_c),
\end{equation}
with the dimensionless response
\begin{equation}
H^{(l_1,l_2,l_3)}(q)
=
4\pi\int_0^1
s^2\psi^{(l_1,l_2,l_3)}(s)
j_{l_2}(qs)\,ds.
\end{equation}
Thus, for a fixed dimensionless radial profile, changing $R_c$ dilates the frequency axis while scaling the response amplitude by $a_{R_c}^{(l_1,l_2,l_3)}R_c^3$. In particular, a characteristic feature at dimensionless frequency $q_0$ appears at $q_{\mathrm{char}}=q_0/R_c$, indicating that smaller cutoffs shift the characteristic frequency scale toward higher frequencies. At low frequencies, the expansion $j_{l_2}(qs)=(qs)^{l_2}/(2l_2+1)!!+O(q^{l_2+2})$ yields $H^{(l_1,l_2,l_3)}(q)=O(q^{l_2})$ as $q\to0$. Nevertheless, oscillations and sidelobes may prevent the response from being a strictly monotonic low-pass filter.

\section{Multi-Cutoff Spectral Aggregation}
\subsection{Radial Spectral Response in MCSA}

Motivated by the cutoff-dependent frequency scaling derived above, we introduce Multi-Cutoff Spectral Aggregation (MCSA) to aggregate radial interactions across multiple spatial scales. Given $N$ cutoff radii $R_c^{(1)}<\cdots<R_c^{(N)}$, let the radial kernel for the $n$-th cutoff be
\begin{equation}
\label{eq:mc_individual_kernel}
\Phi_{n,R_c^{(n)}}^{(l_1,l_2,l_3)}(r)
=\widetilde{\phi}_{n,R_c^{(n)}}^{(l_1,l_2,l_3)}(r)
\mathcal E_{R_c^{(n)}}(r).
\end{equation}
where $\widetilde{\phi}_{n,R_c^{(n)}}^{(l_1,l_2,l_3)}(r)$ denotes the radial basis expansion or learned radial filter associated with the $n$-th cutoff, and $\mathcal E_{R_c^{(n)}}(r)$ is a smooth cutoff envelope that ensures the radial kernel vanishes smoothly at $R_c^{(n)}$. Thus, each kernel defines a radial interaction with its own spatial support and frequency scale. The multi-cutoff message-passing operation is
\begin{equation}
\label{eq:multi_cutoff_mp}
\begin{aligned}
\mathbf x_i'^{(l_3)}
=
\sum_{n=1}^N \gamma_n\sum_{j\in\mathcal N^{(n)}(i)}
\sum_{l_1,l_2}
\mathbf x_j^{(l_1)}
\otimes_{l_1,l_2}^{l_3}
Y^{(l_2)}(\hat{\mathbf r}_{ij})
\Phi_{n,R_c^{(n)}}^{(l_1,l_2,l_3)}(\|\mathbf r_{ij}\|),
\end{aligned}
\end{equation}
where $\gamma_n$ denotes the aggregation weight of the $n$-th cutoff, and $\mathcal N^{(n)}(i)=\{j:|\mathbf r_{ij}|<R_c^{(n)}\}$ is the corresponding neighborhood.  
By the linearity of the Fourier transform, the aggregated radial spectral response satisfies
\begin{equation}
\label{eq:mc_spectral_superposition}
H_{\mathrm{MC}}^{(l_1,l_2,l_3)}(q)
=
\sum_{n=1}^{N}
\gamma_n H_{R_c^{(n)}}^{(l_1,l_2,l_3)}(q).
\end{equation}
Since all branches share the same spherical harmonic basis and Clebsch--Gordan coupling rules, the resulting Fourier-domain operator retains the angular structure of the single-cutoff formulation, with its radial response replaced by $H_{\mathrm{MC}}^{(l_1,l_2,l_3)}(q)$.
To further characterize the resulting spectral response, we
consider a scale-consistent parameterization of each radial branch:
\begin{equation}
\label{eq:mc_scale_consistent}
\Phi_{n,R_c^{(n)}}^{(l_1,l_2,l_3)}(r)
=
a_{R_c^{(n)}}^{(l_1,l_2,l_3)}
\psi_n^{(l_1,l_2,l_3)}
\left(\frac{r}{R_c^{(n)}}\right),
\end{equation}
where $a_{R_c^{(n)}}^{(l_1,l_2,l_3)}$ is a cutoff-dependent
amplitude and $\psi_n^{(l_1,l_2,l_3)}$ is the dimensionless
radial profile of the $n$-th branch, defined on $[0,1]$.
Substituting Eq.~\eqref{eq:mc_scale_consistent} into the
Fourier--Bessel transform and applying the change of variables
$s=r/R_c^{(n)}$ yields
\begin{equation}
\label{eq:mc_individual_scaled_spectrum}
H_{R_c^{(n)}}^{(l_1,l_2,l_3)}(q)
=
a_{R_c^{(n)}}^{(l_1,l_2,l_3)}
\bigl(R_c^{(n)}\bigr)^3
H_n^{(l_1,l_2,l_3)}
\left(qR_c^{(n)}\right),
\end{equation}
where
\begin{equation}
\label{eq:mc_dimensionless_response}
H_n^{(l_1,l_2,l_3)}(q)
=
4\pi\int_0^1
s^2\psi_n^{(l_1,l_2,l_3)}(s)
j_{l_2}(qs)\,ds
\end{equation}
denotes the dimensionless radial frequency response of the
$n$-th branch. Here, $j_{l_2}$ is the spherical Bessel function
of the first kind.

Combining Eqs.~\eqref{eq:mc_spectral_superposition} and
\eqref{eq:mc_individual_scaled_spectrum}, we obtain
\begin{equation}
\label{eq:mc_scaled_spectrum}
\begin{aligned}
H_{\mathrm{MC}}^{(l_1,l_2,l_3)}(q)
=
\sum_{n=1}^{N}
&\gamma_n
a_{R_c^{(n)}}^{(l_1,l_2,l_3)}
\bigl(R_c^{(n)}\bigr)^3
\cdot
H_n^{(l_1,l_2,l_3)}
\left(qR_c^{(n)}\right).
\end{aligned}
\end{equation}
In general, independently parameterized radial branches may
have different dimensionless profiles, resulting in distinct
responses $H_n^{(l_1,l_2,l_3)}$. When all branches share the
same dimensionless radial profile, i.e.,
$\psi_n^{(l_1,l_2,l_3)}=\psi^{(l_1,l_2,l_3)}$ for all $n$,
their dimensionless responses reduce to a common function
$H^{(l_1,l_2,l_3)}$, and Eq.~\eqref{eq:mc_scaled_spectrum}
simplifies to
\begin{equation}
\label{eq:mc_shared_profile_spectrum}
\begin{aligned}
H_{\mathrm{MC}}^{(l_1,l_2,l_3)}(q)
=
\sum_{n=1}^{N}
&\gamma_n
a_{R_c^{(n)}}^{(l_1,l_2,l_3)}
\bigl(R_c^{(n)}\bigr)^3\cdot
H^{(l_1,l_2,l_3)}
\left(qR_c^{(n)}\right).
\end{aligned}
\end{equation}

Thus, under the scale-consistent parameterization, MCSA constructs
a weighted superposition of radially scaled frequency responses,
allowing different spatial scales to contribute to a unified
equivariant interaction. For interatomic potentials, short-range
interactions primarily characterize local bonding and repulsion,
whereas larger neighborhoods provide information about coordination
and the surrounding atomic environment. By aggregating interactions
across these spatial scales, MCSA enriches the radial representation
of the potential energy surface and provides a more flexible basis
for modeling atomic interactions.

\subsection{Computational Cost Reduction via Radial Fusion}
\label{sec:reduce_cost}
A direct implementation may independently evaluate the angular tensor products for every cutoff, introducing redundant computation.
The aggregated radial kernel is defined as
\begin{equation}
\label{eq:multi_cutoff_kernel}
\Phi_{\mathrm{MC}}^{(l_1,l_2,l_3)}(r)
=
\sum_{n=1}^{N}
\gamma_n\Phi_{n,R_c^{(n)}}^{(l_1,l_2,l_3)}(r),
\end{equation}
MCSA avoids this redundancy by exploiting the compact support of the radial kernels and the linearity of message aggregation. Let
$\mathcal N^{(n)}(i)=\{j:r_{ij}<R_c^{(n)}\}$.
Since $\mathcal N^{(n)}(i)\subseteq\mathcal N_{\max}(i)$ and $\Phi_{n,R_c^{(n)}}^{(l_1,l_2,l_3)}(r)=0$ outside its cutoff, we have
\begin{equation}
\label{eq:mc_sum_reordering}
\begin{aligned}
&\sum_{n=1}^{N}\gamma_n
\sum_{j\in\mathcal N^{(n)}(i)}
\mathbf x_j^{(l_1)}
\otimes_{l_1,l_2}^{l_3}
Y^{(l_2)}(\hat{\mathbf r}_{ij})
\Phi_{n,R_c^{(n)}}^{(l_1,l_2,l_3)}(r_{ij})=
\sum_{j\in\mathcal N_{\max}(i)}
\mathbf x_j^{(l_1)}
\otimes_{l_1,l_2}^{l_3}
Y^{(l_2)}(\hat{\mathbf r}_{ij})
\Phi_{\mathrm{MC}}^{(l_1,l_2,l_3)}(r_{ij}).
\end{aligned}
\end{equation}
This identity allows all cutoff-dependent radial contributions to be accumulated before evaluating the CG tensor product. Algorithm~\ref{alg:MCSA} summarizes the resulting computation.

\begin{figure*}[t!] %
  \centering
  \begin{minipage}[t]{0.48\textwidth}
    \centering
    \begin{algorithm}[H] %
      \caption{SO(3)-Equivariant Message Passing}
      \label{alg:egnn}
      \begin{algorithmic} %
      \State Initialize atomic coordinates $coords$
      \State Initialize atomic feature  $\mathbf x$
      \State Initialize Cutoff Radius   $R_c$
      \State Initialize output features $\mathbf{x}' \leftarrow \mathbf{0}$
    \State $E \leftarrow \text{radius\_graph}(coords, R_c)$  
    \For{\textbf{each} $(i, j) \in E$}
        \State $r_{ij} \leftarrow \text{norm}(coords_j - coords_i)$
        \State $Y_{ij} \leftarrow \text{sph\_harm}(coords_j - coords_i)$
        \State $\Phi_{ij} \leftarrow \text{radial\_weight}(r_{ij},R_c)$ 
        \State $m_{ij} \leftarrow \text{weighted\_TP}(\mathbf x_j, Y_{ij},\Phi_{ij})$ 
     \State $\mathbf x'_i \leftarrow \mathbf x'_i+m_{ij}$
    \EndFor
      \end{algorithmic}
    \end{algorithm}
  \end{minipage}
  \hfill %
  \begin{minipage}[t]{0.48\textwidth}
    \centering
    \begin{algorithm}[H]
      \caption{SO(3)-Equivariant Message Passing with MCSA}
      \label{alg:MCSA}
      \begin{algorithmic}
         \State $E_\text{max} \leftarrow \text{radius\_graph}(coords, \max(R^{n}_c))$  
    \For{\textbf{each} $(i, j) \in E_\text{max}$}
        \State $r_{ij} \leftarrow \text{norm}(coords_j - coords_i)$
        \State $Y_{ij} \leftarrow \text{sph\_harm}(coords_j - coords_i)$
        \State $\Phi^\mathrm{MC}_{ij}\gets 0$
        \For{\textbf{each} $n = 1:N$}
        \State $\Phi^{MC}_{ij} \leftarrow \Phi^{MC}_{ij}+\gamma_n\cdot\text{radial\_weight}(r_{ij},R_c^n)$ 
        \EndFor
        \State $m_{ij} \leftarrow \text{weighted\_TP}(\mathbf x_j, Y_{ij},\Phi^{MC}_{ij})$ 
        \State $\mathbf x'_i \leftarrow \mathbf x'_i+m_{ij}$
    \EndFor
      \end{algorithmic}
    \end{algorithm}
  \end{minipage}
  \label{alg:both}
\end{figure*}

As shown in Algorithm~\ref{alg:MCSA}, each cutoff contributes only to the scalar radial weight, while the CG tensor product is evaluated once per edge and coupling path, independently of $N$. For a direct implementation that independently evaluates
the CG tensor products for all cutoff branches, the total
number of angular contractions scales as
$\sum_{n=1}^{N}|\mathcal E_n|$.
By exploiting the shared angular basis and the linearity
of tensor-product operations, radial fusion reduces this
to $|\mathcal E_{\max}|$, requiring only one angular
contraction per edge in the maximum neighborhood.
The cutoff-dependent radial filters must still be evaluated
within their respective interaction ranges, so their
computational cost remains dependent on
$\sum_{n=1}^{N}|\mathcal E_n|$.
Consequently, MCSA reduces redundant angular computations,
while the overall runtime savings depend on the relative
costs of radial evaluations and angular contractions,
as well as the neighborhood sizes.

\subsection{Plug-and-Play Integration}

MCSA modifies only the invariant scalar radial weights, leaving the spherical harmonics and CG coupling structure unchanged. It can therefore be incorporated into existing SO(3)-equivariant message-passing layers whose edge interactions admit a separable radial-weighted form and linear aggregation, including compatible radial interaction blocks in architectures such as NequIP, MACE, EquiformerV2, and eSCN. For attention-based layers, exact fusion additionally requires preserving any cutoff-dependent masks and nonlinear attention normalization.

The mechanism requires no change to the angular representation or its coupling rules. Moreover, when $N=1$ and $\gamma_1=1$, MCSA recovers the corresponding single-cutoff radial kernel and hence the original message-passing formulation. Since a finite weighted sum of sufficiently smooth, compactly supported radial kernels retains the required cutoff smoothness, MCSA also preserves the differentiability conditions needed for energy and force prediction.

\section{Experiments}
\label{sec:experiments}

We evaluate MCSA on multiple  benchmarks to demonstrate its effectiveness in improving geometric representation fidelity, higher-order derivative accuracy, and long-term simulation stability. Unless otherwise specified, the EGNN+MCSA
configuration uses $N=3$ with cutoff radii
$\mathcal R_{\rm multi}=[2.0,4.0,6.0]$ \AA,
incorporating the Radial Fusion optimization.

\subsection{Generality Across Backbones}

To verify the universality of the Multi-Cutoff Spectral Aggregation (MCSA) module, we integrate it into representative SO(3)-equivariant GNNs: Nequix~\cite{koker2025training}, Eqnorm~\cite{eqnorm2025}, and EquiformerV3~\cite{equiformer_v3}. We evaluate three representative SO(3)-equivariant
backbones on the MPtrj dataset \cite{jain2013commentary, deng2023chgnet} using matched training
settings for their baseline and MCSA variants. As shown in Table~\ref{tab:generality}, we evaluate performance on both the in-distribution (ID) MPtrj test split and the out-of-distribution (OOD) Alexandria dataset\cite{schmidt2023machine, schmidt2021crystal}.

\begin{table}[htbp]
\centering
\caption{Performance improvements of EGNNs with MCSA module on two datasets. Mean Absolute Error (MAE): energy (meV/atom), force (meV/\AA), stress (meV/atom/\AA$^3$).}
\label{tab:generality}
\resizebox{0.8\textwidth}{!}{
\begin{tabular}{lccc|ccc}
\toprule
\multirow{2}{*}{Method} & \multicolumn{3}{c}{MPtrj (ID)} & \multicolumn{3}{c}{Alexandria (OOD)} \\
\cmidrule(lr){2-4} \cmidrule(lr){5-7}
 & Energy & Force & Stress & Energy & Force & Stress \\
\midrule
Nequix     & 14.2 & 40.4 & 0.273 & 41.8  & 75.5  & 0.366  \\
\rowcolor{gray!10}
Nequix+MCSA         & 13.4 & 38.7 & 0.267 &  39.9 & 74.3 & 0.354 \\

Eqnorm        & 11.5 & 34.2 & 0.211 & 36.7 & 68.3 & 0.331 \\
\rowcolor{gray!10}
Eqnorm+MCSA           & 10.8 & 32.5 & 0.203 & 35.6 & 67.1 & 0.323 \\

EquiformerV3 & 6.3 & 18.1 & 0.151 & 22.1 & 43.5 & 0.217 \\
\rowcolor{gray!10}
EquiformerV3+MCSA    & \textbf{6.2} & \textbf{17.2} & \textbf{0.147} & \textbf{20.3} &   \textbf{41.1} & \textbf{0.206}  \\
\bottomrule
\end{tabular}
}
\end{table}
As shown in Table~\ref{tab:generality}, MCSA consistently improves energy, force, and stress prediction across all three architectures on both ID and OOD datasets. Notably, the improvements extend beyond the training distribution, indicating better generalization to unseen atomic structures. For EquiformerV3, MCSA reduces OOD energy, force, and stress MAEs by 8.1\%, 5.5\%, and 5.1\%, respectively. Similar improvements are observed in Nequix and Eqnorm, demonstrating that the benefits of multi-cutoff radial aggregation are not restricted to a specific backbone. These results highlight the effectiveness and transferability of MCSA as a lightweight enhancement for improving the radial representation of equivariant interatomic potentials.

\subsection{Performance on Higher-Order Derivatives}
While fitting energy and forces (first-order derivatives) is the primary objective of MLIPs, the true reliability of a learned potential energy surface is governed by its higher-order derivatives. 

\subsubsection{Phonon Dispersion Analysis}

\begin{figure*}[t]  
    \centering
    \begin{subfigure}[b]{0.32\textwidth}
        \centering
        \includegraphics[width=\linewidth]{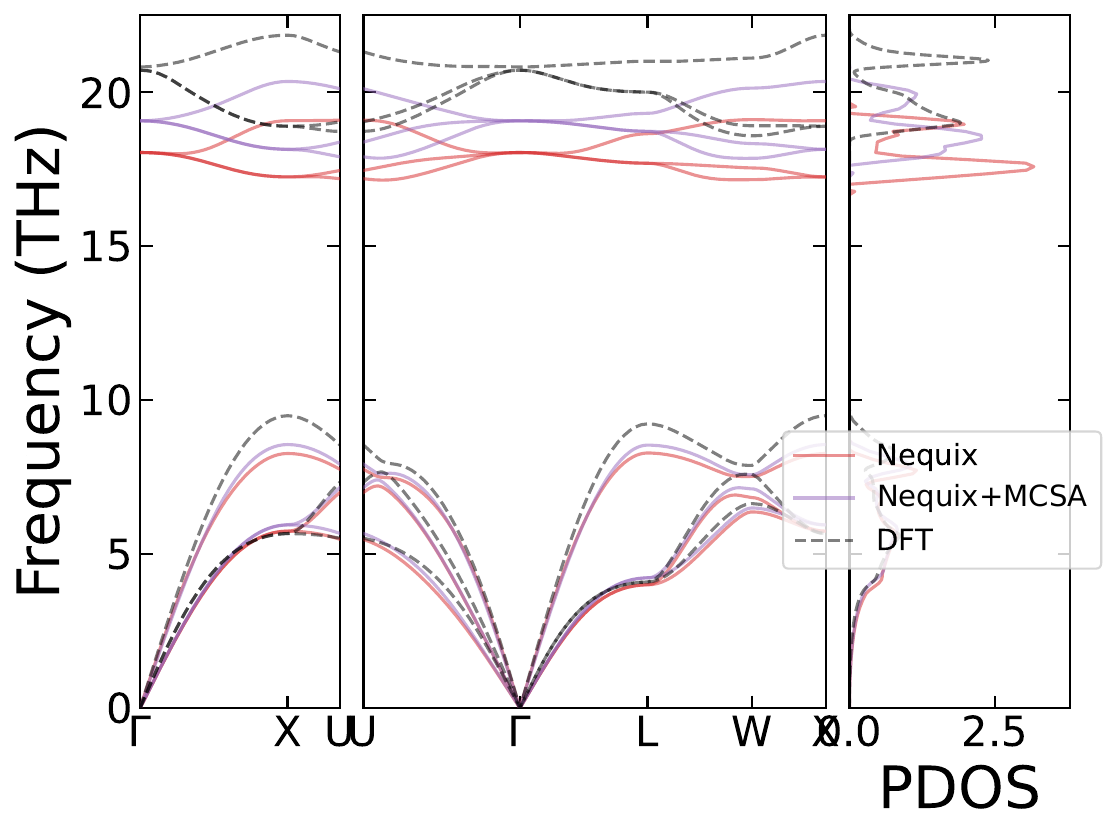}
        \caption{}
        \label{fig:phonon_A}
    \end{subfigure}
    \hfill
    \begin{subfigure}[b]{0.32\textwidth}
        \centering
        \includegraphics[width=\linewidth]{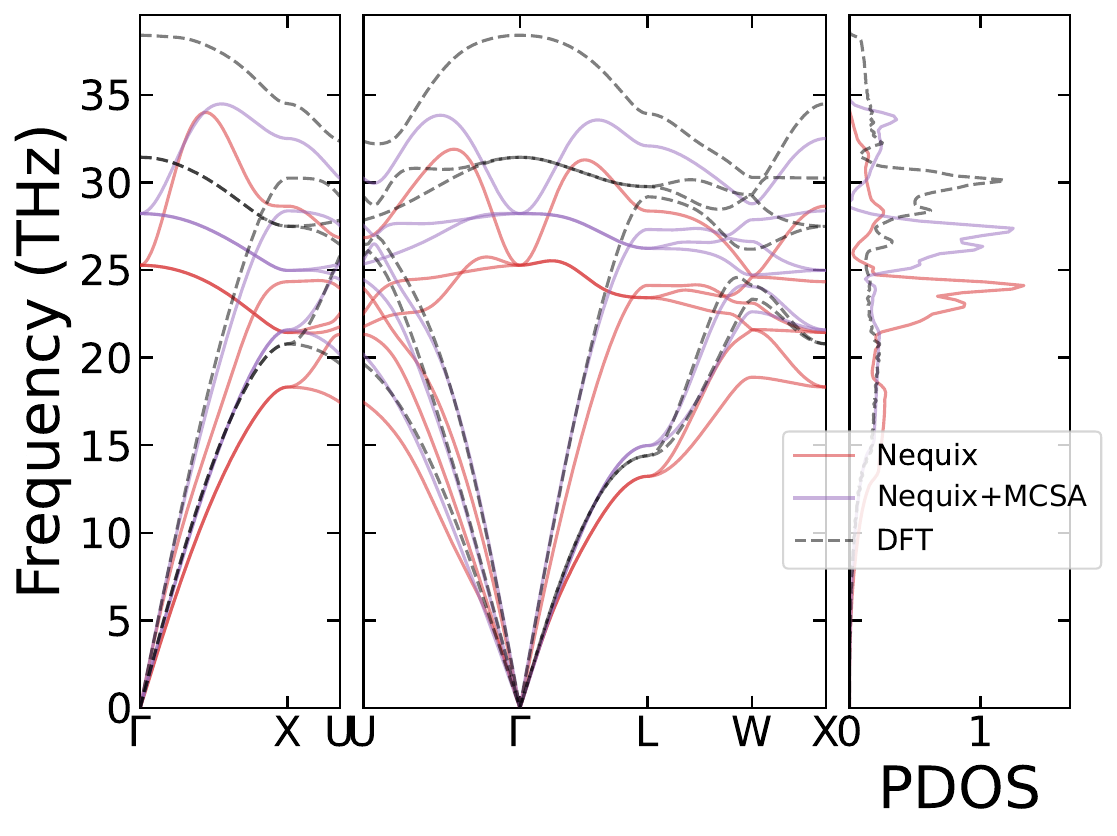}
        \caption{}
        \label{fig:phonon_B}
    \end{subfigure}
    \hfill
    \begin{subfigure}[b]{0.32\textwidth}
        \centering
        \includegraphics[width=\linewidth]{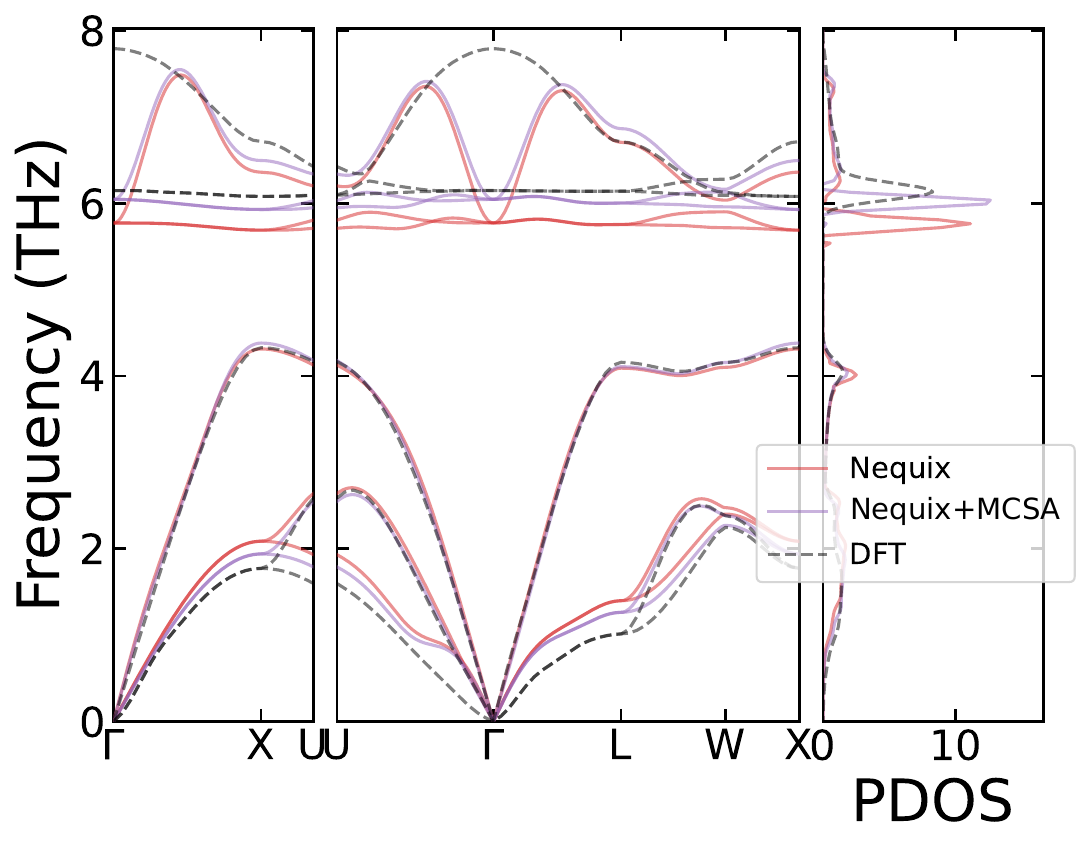}
        \caption{}
        \label{fig:phonon_C}
    \end{subfigure}
    \caption{Comparison of phonon spectra and phonon density of states (PDOS) inferred by different MLIPs against DFT for three materials.
    (a) BAs. (b) BN. (c) CaSe.}
    \label{fig:phonon_compare_three_materials}
\end{figure*}

We evaluate the fidelity of the learned potential energy surfaces (PESs) through phonon dispersion relations and phonon density of states (PDOS), which provide a stringent assessment of their local curvature. Specifically, squared phonon frequencies correspond to the eigenvalues
of the mass-weighted dynamical matrix, which is constructed
from the interatomic force constants \cite{TOGO20151}. It makes phonon properties particularly sensitive to errors in the second-order derivatives of the PES. Figure~\ref{fig:phonon_compare_three_materials} compares the predicted phonon spectra of three representative materials: BAs, BN, and CaSe. The baseline Nequix exhibits noticeable deviations from DFT references, particularly in the high-frequency optical branches. This discrepancy is especially pronounced in strongly covalent BN, whose phonon modes extend to approximately 38 THz. The baseline underestimates the highest optical frequencies by nearly 10 THz and produces a substantially distorted PDOS, indicating insufficient accuracy in capturing the local curvature associated with stiff atomic bonds.

In contrast, Nequix+MCSA achieves closer agreement with DFT across all three materials, improving the prediction of both acoustic and optical branches. In particular, the upper optical branches are shifted toward the reference frequencies, leading to better alignment with the corresponding PDOS peaks. By combining radial interactions across multiple cutoff scales, MCSA enriches the representation of local atomic geometries while retaining information from broader atomic environments. These improvements suggest that multi-cutoff radial aggregation provides a more expressive representation for capturing the fine-scale structural variations underlying higher-order physical properties.

\subsubsection{Lattice Thermal Conductivity}

Accurate lattice thermal conductivity prediction requires
reliable second-order harmonic and third-order anharmonic
interatomic force constants, while fourth-order force
constants are additionally required when four-phonon
scattering is considered.\cite{hellman2014phonon, lindsay2019perspective, li2014shengbte}

As shown in Table~\ref{tab:thermal}, we evaluate the prediction performance on the PhononDB-103 thermal conductivity benchmark \cite{pta2024thermalconductivitypredictionsfoundation}. We report the Symmetric Relative Mean Error ($\kappa_{SRME}$) \cite{pta2024thermalconductivitypredictionsfoundation} to account for the logarithmic distribution of $\kappa$ across different materials. To directly validate the efficiency of MCSA, we also record the average computational time required to process each material.

\begin{table}[htbp]
\centering
\caption{Lattice thermal conductivity ($\kappa$) prediction performance and computational efficiency on the PhononDB-103 dataset. A lower $\kappa_{SRME}$ indicates better agreement with ab initio references. The inference time is measured in minutes per material.}
\label{tab:thermal}
\begin{tabular}{lcc}
\toprule
Model & Avg. Time (min/mat) & $\kappa_{SRME}$ ($\downarrow$) \\
\midrule
Nequix     & 1.107  & 0.487 \\
Nequix+MCSA & 1.136  & \textbf{0.420} \\
\bottomrule
\end{tabular}
\end{table}
The results demonstrate that the integration of the MCSA module reduces the $\kappa_{SRME}$ from 0.487 to 0.420, representing a significant 13.7\% relative reduction in error. Remarkably, this enhancement in capturing high-order anharmonicity comes at an almost negligible computational cost. The average calculation time per material increases by merely $\sim$1.7 seconds (from 1.107 to 1.136 minutes), an overhead of less than 3\%. These results demonstrate that MCSA improves lattice
thermal conductivity prediction with only a modest
increase in computational time, consistent with the
efficiency of the proposed radial fusion formulation.

\subsection{PES Smoothness and Molecular Dynamics Stability}
\label{sec:smoothness_and_stability}

Beyond static prediction accuracy, a reliable MLIP should maintain a smooth potential energy surface (PES) and stable long-term molecular dynamics (MD) trajectories. We evaluate these properties through a Si-Si dissociation process and long-time MD simulations.

\begin{figure*}[t]
    \centering
    \begin{subfigure}[b]{0.32\textwidth}
        \centering
        \includegraphics[width=\linewidth]{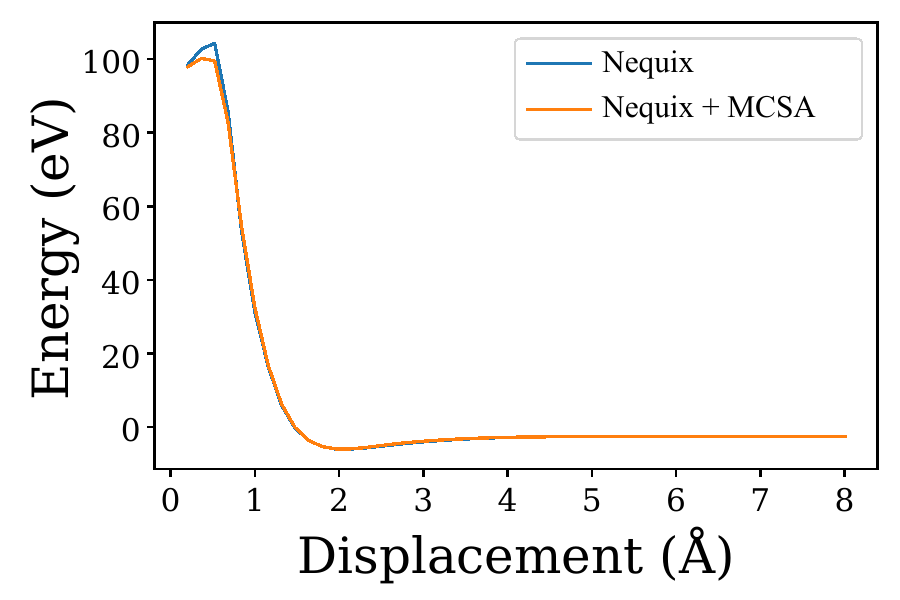}
        \caption{}
        \label{fig:energy_si-si}
    \end{subfigure}
    \hfill
    \begin{subfigure}[b]{0.32\textwidth}
        \centering
        \includegraphics[width=\linewidth]{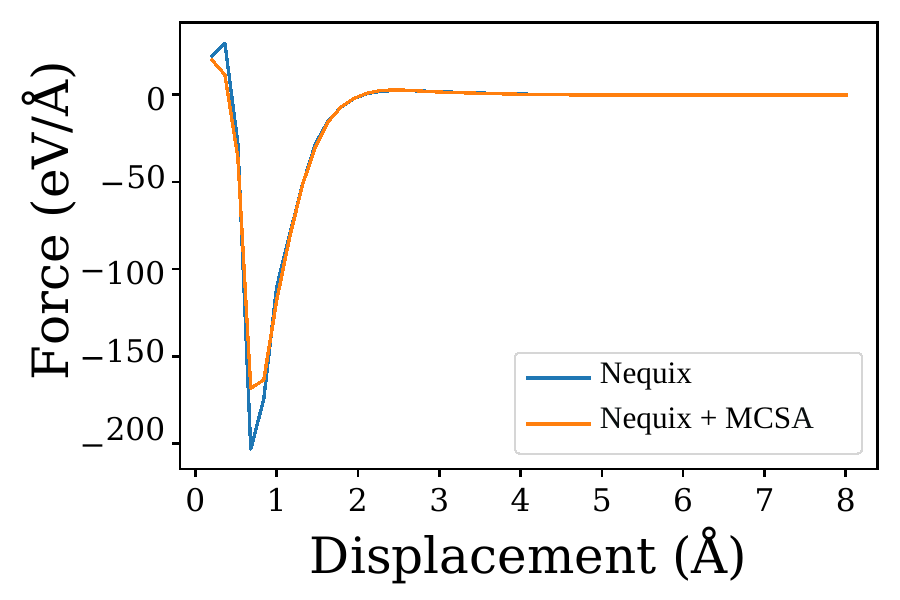}
        \caption{}
        \label{fig:force_si-si}
    \end{subfigure}
    \hfill
    \begin{subfigure}[b]{0.32\textwidth}
        \centering
        \includegraphics[width=\linewidth]{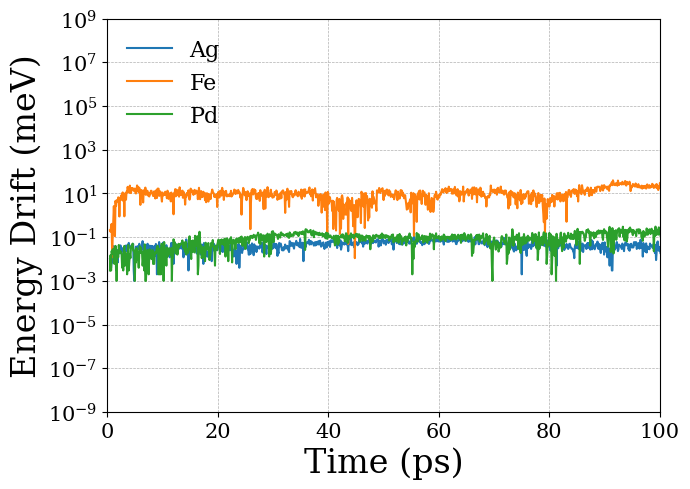}
        \caption{}
        \label{fig:md_tm23}
    \end{subfigure}
    \caption{
    PES smoothness and MD stability.
    (a) Predicted potential energy and (b) interatomic force during Si-Si dissociation using Nequix+MCSA ($N=3$, $\mathcal R_{\rm multi}=[2,4,6]$ \AA).
    (c) Energy evolution during a 100 ps NVE simulation on TM23 at 300 K using Nequix+MCSA.
    }
    \label{fig:smoothness_md}
\end{figure*}

Figures~\ref{fig:smoothness_md}(a) and (b) show that Nequix+MCSA maintains smooth energy and force profiles across the intermediate cutoff boundaries at $2.0$ and $4.0$ \AA, without visible discontinuities or abrupt transitions. This behavior follows from the compactly supported smooth envelopes: their weighted sum preserves the required cutoff smoothness of the individual radial kernels, avoiding artificial artifacts introduced by multi-cutoff aggregation.

We further evaluate dynamical stability through 100 ps NVE simulations on the TM23 dataset \cite{owen2024complexity} at 300 K with a 0.5 fs time step. As shown in Figure~\ref{fig:smoothness_md}(c), Nequix+MCSA completes the entire trajectory without structural collapse, demonstrating stable MD behavior under the tested conditions.

\subsection{Physically Motivated Multi-Cutoff Design for Layered Materials}

To evaluate MCSA on layered materials with distinct interaction length scales, we consider three representative Mo--S structures from JARVIS-DFT \cite{choudhary2020jarvis}: JVASP-100856, JVASP-28733, and JVASP-34141. These structures are treated as held-out test cases. Before model training, all ALIGNN-FF-DB configurations corresponding to these three JARVIS entries were excluded from the training and validation splits. 
Structural analysis reveals two characteristic interaction scales: in-plane Mo--S covalent bonds with distances around $2.4 $ to $ 3.0$ \AA, and interlayer S$-$S separations ranging from $3.61$ to $6.51$ \AA. Based on these structural characteristics, we adopt a physically motivated dual-cutoff configuration of $[3.0,7.0]$ \AA. The shorter cutoff emphasizes local in-plane
environments, while the larger cutoff extends the
interaction range to incorporate neighboring layers.

\begin{figure*}[t]
    \centering
        \includegraphics[
            width=\linewidth,
        ]{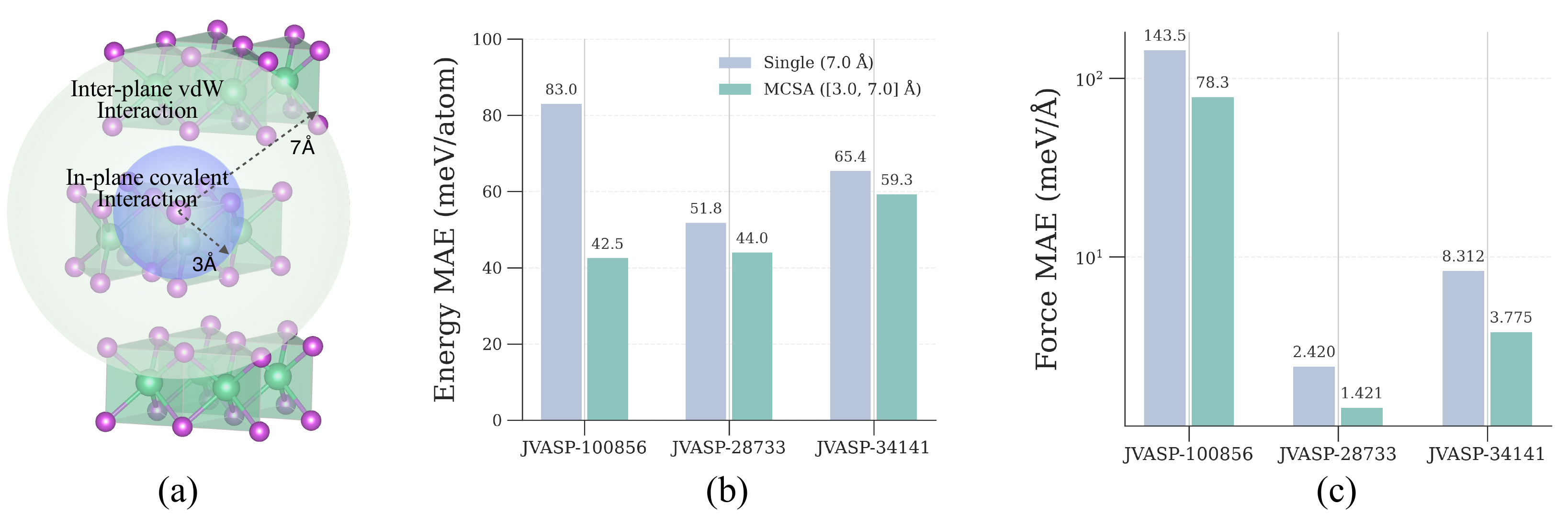}
        
\caption{
(a) Mo--S crystal structure illustrating the multi-cutoff design for intra- and interlayer interactions. 
        (b) Energy MAE (meV/atom) and (c) Force MAE (meV/\AA) in log scale on three representative layered Mo--S structures, comparing single-cutoff and MCSA on Nequix.}
        \label{fig:layered_mos}
\end{figure*}

We train two models on the ALIGNN-FF-DB dataset \cite{alignn} using identical architectures and training configurations, differing only in their cutoff settings: a single-cutoff baseline with $R_c=7.0$ \AA, and MCSA with $R_c=[3.0,7.0]$ \AA. Figure~\ref{fig:layered_mos} reports their energy and force prediction errors on the three representative structures. MCSA consistently reduces both energy and force MAEs across all three systems, with particularly notable improvements in force prediction. These results demonstrate the potential of physically motivated multi-cutoff aggregation for modeling materials with distinct short-range and interlayer interaction scales.

\subsection{Ablation on MCSA Configurations}
\label{sec:ablation_radial}

We conduct comprehensive ablation studies to investigate the effects of three key MCSA configurations: the number and placement of cutoff radii ($\mathcal R_{\rm multi}$), the radial basis function (RBF) allocation, and the polynomial envelope degree ($p$). We additionally report the average inference latency per material, measured in milliseconds on a single NVIDIA A800 GPU, to assess the accuracy-efficiency trade-off. All models are trained on MPtrj and evaluated on its test split. The results are summarized in Table~\ref{tab:comprehensive_ablation}.

\begin{table*}[htbp]
\centering
\caption{Comprehensive ablation study on radial configurations using the Nequix baseline. MAE is reported in meV/atom (Energy), meV/Å (Force), and meV/atom/Å$^3$ (Stress). The inference time demonstrates the marginal overhead of the MCSA mechanism.}
\label{tab:comprehensive_ablation}
\resizebox{\textwidth}{!}{
\begin{tabular}{lcccccc|ccc}
\toprule
\multirow{2}{*}{Variant} & \multirow{2}{*}{$N$} & \multirow{2}{*}{Cutoff Radii $\mathcal{R}_{\rm multi}$ (\AA)} & \multirow{2}{*}{RBF Size} & \multirow{2}{*}{Poly Degree $p$} & \multirow{2}{*}{Inference Time} & \multirow{2}{*}{Params} & \multicolumn{3}{c}{MAE ($\downarrow$)} \\
\cmidrule(lr){8-10}
 & & & & & & & Energy & Force & Stress \\
\midrule
Baseline & 1 & $[6.0]$ & $[8]$ & $[6]$ & 47 ms & 0.7M & 14.15 & 40.41 & 0.273 \\
\midrule
\multicolumn{10}{l}{\textit{Part I: Number and Placement of Cutoff Radii}}  \\
Two-scale & 2 & $[2.0, 6.0]$ & $[8, 8]$ & $[6, 6]$ & 50 ms & 0.9M & 13.70 & 39.25 & 0.268 \\
\rowcolor{gray!10} Three-scale (Ours) & 3 & $[2.0, 4.0, 6.0]$ & $[8, 8, 8]$ & $[6, 6, 6]$ & 51 ms & 1.1M & \textbf{13.44} & \textbf{38.54} & \textbf{0.267} \\
Four-scale (A) & 4 & $[3.0, 4.0, 5.0, 6.0]$ & $[8, \times 4]$ & $[6, \times 4]$ & 54 ms & 1.3M & 13.57 & 38.87 & 0.269 \\
Four-scale (B) & 4 & $[1.0, 2.0, 3.0, 6.0]$ & $[8, \times 4]$ & $[6, \times 4]$ & 54 ms & 1.3M & 13.79 & 39.34 & 0.268 \\
Six-scale & 6 & $[1.0, 2.0, 3.0, 4.0, 5.0, 6.0]$ & $[8, \times 6]$ & $[6, \times 6]$ & 56 ms & 1.6M & 14.26 & 40.18 & 0.276 \\
\midrule
\multicolumn{10}{l}{\textit{Part II: Radial Basis and Envelope Configurations}} \\
Outer-Heavy RBF & 2 & $[3.0, 6.0]$ & $[8, 16]$ & $[6, 6]$ & 52 ms & 0.9M & 13.76 & 39.28 & \textbf{0.267} \\
Inner-Heavy RBF & 2 & $[3.0, 6.0]$ & $[16, 8]$ & $[6, 6]$ & 52 ms & 0.9M & 13.68 & 39.15 & 0.272 \\
Envelope Mix (A) & 2 & $[3.0, 6.0]$ & $[8, 8]$ & $[6, 8]$ & 50 ms & 0.9M & 13.83 & 39.71 & 0.270 \\
Envelope Mix (B) & 2 & $[3.0, 6.0]$ & $[8, 8]$ & $[8, 6]$ & 49 ms & 0.9M & 13.88 & 39.74 & 0.269 \\
\bottomrule
\end{tabular}
}
\end{table*}
Part I of Table~\ref{tab:comprehensive_ablation} shows that three cutoffs with $\mathcal R_{\rm multi}=[2.0,4.0,6.0]$ \AA\ achieve the best overall performance, reducing Energy, Force, and Stress MAEs to 13.44, 38.54, and 0.267, respectively. Increasing $N$ beyond three brings no further gains, indicating that selecting complementary interaction scales is more important than simply adding cutoffs.

MCSA introduces only modest inference overhead, increasing latency from 47 ms to 51 ms for $N=3$ (8.5\%). This demonstrates the efficiency of radial fusion, which avoids redundant angular tensor-product computations across cutoff branches.

Part II shows that the Inner-Heavy RBF allocation ($[16,8]$) improves Energy and Force accuracy over the Outer-Heavy configuration ($[8,16]$), while slightly increasing Stress error. The envelope configurations yield comparable results, suggesting that cutoff selection has a greater impact than the tested envelope-degree variations.

\section{Related Works}

\textbf{Machine Learning Interatomic Potentials (MLIPs).} The evolution of MLIPs has transitioned from hand-crafted descriptors like Behler-Parrinello NNs \cite{behler2007generalized}, GAP \cite{bartok2010gaussian}, and ACE \cite{drautz2019atomic} to end-to-end Graph Neural Networks (GNNs). Early GNNs, such as SchNet \cite{schutt2018schnet}
and DimeNet \cite{gasteiger2020dimenet}, incorporated
geometric information through continuous-filter
convolutions and directional message passing,
primarily using invariant scalar features. The emergence of E(3)-equivariant architectures \cite{pmlr-v139-schutt21a} marked a shift toward strictly enforcing symmetry on tensor features. NequIP \cite{Batzner2022} established the high data efficiency of equivariant spherical harmonics, leading to diverse architectures: Allegro \cite{Musaelian2023Allegro} prioritized strict locality via many-body tensor products, while MACE \cite{Batatia2022mace} unified ACE and GNNs through body-ordered symmetric contractions. Recent state-of-the-art models like EquiformerV3 \cite{equiformer_v3}, eSCN \cite{escn} and eSEN \cite{esen} further optimize these concepts via attention mechanisms and efficient tensor operations.
 
\textbf{Radial Representations, Cutoff Design, and Multi-Scale Interactions.}
Radial basis representations are fundamental to MLIPs, ranging from atom-centered symmetry functions \cite{behler2007generalized} and SOAP \cite{bartok2010gaussian,bartok2013soap} to MTP \cite{shapeev2016mtp} and ACE \cite{drautz2019atomic}. Subsequent studies have explored more efficient radial basis parameterizations and expansions \cite{goscinski2021optimal,bochkarev2022ace}. In neural architectures, SchNet \cite{schutt2018schnet} and DimeNet \cite{gasteiger2020dimenet} employ continuous radial filters, while equivariant models such as NequIP \cite{Batzner2022}, MACE \cite{Batatia2022mace}, and Allegro \cite{Musaelian2023Allegro} integrate learnable radial functions with smooth cutoff envelopes. Recent efforts further investigate equivariant architecture design and interaction ranges through many-body representations, nonlocal interactions, and adaptive cutoff strategies \cite{batatia2025designspace,chen2022m3gnet,unke2021spookynet,lin2023complete,ji2025sognet,han2026dynamic}. Unlike approaches that primarily modify radial parameterization or neighborhood selection, MCSA focuses on the spectral properties of radial kernels, combining differently scaled cutoff responses to enrich multi-scale representations while retaining the original equivariant tensor-product structure.
 
\section{Conclusion}
In this work, we proposed Multi-Cutoff Spectral Aggregation
(MCSA), a plug-and-play mechanism that enriches the radial
representations of SO(3)-equivariant GNNs through the
superposition of differently scaled cutoff responses.
By fusing cutoff-dependent radial weights before angular
tensor-product operations, MCSA enables efficient
multi-scale interactions while preserving rotational
equivariance. Extensive experiments demonstrate
improvements in energy, force, and stress prediction,
as well as higher-order properties including phonon
dispersion and lattice thermal conductivity. Further
evaluations on layered materials and long-term molecular
dynamics demonstrate the effectiveness and stability
of the proposed approach.
\bibliography{main}
\bibliographystyle{unsrt}

\newpage
\appendix
\section{Limitations}
\label{appx:Limitations}
Despite its efficacy, MCSA currently relies on predefined radial scales, which may not be universally optimal for all chemical environments. Exploring adaptive cutoff scheduling remains a promising direction. Additionally, while our associative optimization ensures efficiency by fusing scales linearly in the scalar domain, future work could investigate lightweight cross-scale interaction modules to capture potential non-linear couplings between different spatial resolutions.

\section{Implementation Details}
\label{appx:details}

We implemented the Multi-Cutoff Spectral Aggregation (MCSA) strategy on three SO(3)-equivariant architectures: Nequix\cite{koker2025training}, EquiformerV3\cite{equiformer_v3}, and Eqnorm\cite{eqnorm2025}. All models were trained using multi-cutoff radii $\mathcal{R}_{\rm multi} = \{2.0, 4.0, 6.0\}$ Å.

\subsection{Nequix-MCSA}
We train the Nequix-MCSA, following the training protocol of Nequix \cite{koker2025training}, jointly optimizing the Mean Absolute Error (MAE) for energy, forces, and macroscopic stress. To efficiently accelerate convergence, we employ the Muon optimizer. Unlike the EquiformerV3 \cite{equiformer_v3} setup, no auxiliary positional denoising tasks were utilized during the training of this backbone. The complete set of training hyper-parameters is detailed in Table~\ref{appendix:tab:nequix_hyperparameters}. The model was trained across 120 NVIDIA A800 GPUs, and the entire training process takes approximately 1617 GPU-hours.

\begin{table}[htp]
\centering
\caption{Hyper-parameters for Nequix-MCSA.}
\scalebox{0.75}{
\begin{tabular}{ll}
\toprule[1.2pt]
Hyper-parameters & Value \\
\midrule[1.2pt]
Optimizer & Muon \\
Learning rate scheduling & Cosine decay + linear warmup \\
Warmup epochs & $0.1$ \\
Maximum learning rate & $1 \times 10^{-2}$ \\
Batch size (per GPU) & $64$ \\
Weight decay & $1 \times 10 ^{-3}$ \\ 
Energy coefficient $\lambda_{E}$ & $20$ \\
Force coefficient $\lambda_{F}$ & $20$ \\
Stress coefficient $\lambda_s$ & $5$ \\
Gradient clipping norm threshold & $100$ \\
Model EMA decay & $0.999$ \\
\midrule[0.6pt]
\textcolor{blue}{Cutoffs (\si{\angstrom})}  & $[2,4,6]$ \\
Radial basis function & Bessel \\
\textcolor{blue}{Number of radial bases} & $[8, 8, 8]$ \\
\textcolor{blue}{Radial polynomial envelope degree $p$} & $[6.0, 6.0, 6.0]$ \\
Radial MLP hidden dimension & $64$ \\
Radial MLP layers & $2$ \\
\midrule[0.6pt]
Maximum degree $L_{\text{max}}$ & $3$ \\
Number of message passing layers & $4$ \\
Hidden Irreps & $128\times0e + 64\times1o + 32\times2e + 32\times3o$ \\
\bottomrule[1.2pt]
\end{tabular}
}
\vspace{1mm}
\label{appendix:tab:nequix_hyperparameters}
\end{table}

\subsection{EquiformerV3-MCSA}

We follow the practice of direct pre-training and gradient fine-tuning \cite{equiformer_v3}. Auxiliary task of denoising based on DeNS \cite{dens} was used in  direct pre-training. For DeNS, after sampling which atoms to corrupt, we ensure that the fraction of corrupted atoms does not exceed 0.75, preventing denoising from degenerating into the case where noise is added for all atoms. The hyper-parameters are in Table \ref{appendix:tab:eqv3-MCSA}. 120 A800 GPUs are used for training. Direct pre-training takes about 1680 GPU-hours, and
gradient fine-tuning takes about 961 GPU-hours.

\begin{table}[htp]
\centering
\caption{Hyper-parameters for EquiformerV3-MCSA.}
\scalebox{0.75}{
\begin{tabular}{lll}
\toprule[1.2pt]
Hyper-parameters & Direct pre-training & Gradient fine-tuning \\
\midrule[1.2pt]
Optimizer & AdamW & AdamW \\
Learning rate scheduling & Cosine decay + linear warmup & Cosine decay + linear warmup \\
Warmup epochs & $0.1$ & $0.1$ \\
Maximum learning rate & $2 \times 10 ^{-4}$ & $5 \times 10 ^{-5}$ \\
Batch size (per GPU) & $16$ & $16$ \\
Number of epochs & $70$ & $10$ \\
Weight decay & $1 \times 10 ^{-3}$ & $1 \times 10 ^{-3}$ \\ 
Dropout rate & $0.1$ & $0.0$ \\
Stochastic depth & $0.05$ & $0.0$ \\
Energy coefficient $\lambda_{E}$ & $20$ & $20$ \\
Force coefficient $\lambda_{F}$ & $20$ & $20$ \\
Stress coefficient $\lambda_s$ & $5$ & $5$ \\
Gradient clipping norm threshold & $100$ & $100$ \\
Model EMA decay & $0.999$ & $0.999$ \\
\midrule[0.6pt]
\textcolor{blue}{Cutoffs (\si{\angstrom})}  & $[2,4,6]$ & $[2,4,6]$ \\
Maximum number of neighbors & $300$ & $300$ \\
Radial basis function & Gaussian & Gaussian \\
Number of radial bases & $10$ & $10$ \\
Dimension of hidden scalar features in radial functions $d_{\text{edge}}$ & $(0, 128)$ & $(0, 128)$ \\
\midrule[0.6pt]
Maximum degree $L_{\text{max}}$ & $4$ & $4$ \\
Maximum order $M_{\text{max}}$ & $2$ & $2$ \\
Number of Transformer blocks & $7$ & $7$ \\
Embedding dimension $d_{\text{embed}}$ & $(4, 128)$ & $(4, 128)$ \\ 
$f_{ij}^{(L)}$ dimension after SwiGLU-$S^2$ $d_{\text{attn\_hidden}}$  & $(4, 32)$ & $(4, 32)$ \\
Number of attention heads $h$ & $8$ & $8$\\
$f_{ij}^{(0)}$ dimension $d_{\text{attn\_alpha}}$ & $(0, 64)$ & $(0, 64)$ \\
Value dimension $d_{\text{attn\_value}}$ & $(4, 16)$ & $(4, 16)$ \\
Hidden dimension in feedforward networks after SwiGLU-$S^2$ $d_{\text{ffn}}$ & $(4, 512)$ & $(4, 512)$ \\
Resolution of grid points in attention $(R_{\phi}, R_{\theta})$ & $(8, 14)$ & $(8, 14)$ \\
Resolution of grid points in feedforward networks $(R_{\phi}, R_{\theta})$ & $(14, 14)$ & $(14, 14)$ \\
\midrule[0.6pt]
Probability of optimizing DeNS $p_{dens}$ & $0.5$ & $0.0$ \\
DeNS coefficient $\lambda_dens$ & $10$ & $0.0$ \\
Standard deviation of Gaussian noise $\sigma$ & $0.025$ & $0.0$ \\
Corruption ratio $r_{{dens}}$ & $0.5$ & $0.0$ \\ 
\bottomrule[1.2pt]
\end{tabular}
}
\vspace{1mm}
\label{appendix:tab:eqv3-MCSA}
\end{table}

\subsection{Eqnorm-MCSA}
Similar to the Nequix backbone, we train the Eqnorm-MCSA model end-to-end to jointly optimize the Mean Absolute Error (MAE) for energy, forces, and macroscopic stress. To efficiently process the stress tensor derivations, the model utilizes an edge-based gradient computation mode. We employ the Muon optimizer to accelerate convergence without relying on auxiliary denoising tasks. The comprehensive hyper-parameters are summarized in Table~\ref{appendix:tab:eqnorm_hyperparameters}. The model was trained across 120 NVIDIA A800 GPUs, and the entire training process takes approximately about 2341 GPU-hours.

\begin{table}[htp]
\centering
\caption{Hyper-parameters for Eqnorm-MCSA.}
\scalebox{0.75}{
\begin{tabular}{ll}
\toprule[1.2pt]
Hyper-parameters & Value \\
\midrule[1.2pt]
Optimizer & Muon \\
Learning rate scheduling & Cosine decay + linear warmup \\
Warmup epochs & $0.1$ \\
Maximum learning rate & $1 \times 10^{-2}$ \\
Batch size (per GPU) & $8$ \\
Weight decay & $1 \times 10 ^{-3}$ \\ 
Energy coefficient $\lambda_{E}$ & $20$ \\
Force coefficient $\lambda_{F}$ & $20$ \\
Stress coefficient $\lambda_s$ & $5$ \\
Gradient clipping norm threshold & $100$ \\
Model EMA decay & $0.999$ \\
\midrule[0.6pt]
\textcolor{blue}{Cutoffs (\si{\angstrom})}  & $[2,4,6]$ \\
Radial basis function & Bessel \\
\textcolor{blue}{Number of radial bases} & $[8, 8, 8]$ \\
\textcolor{blue}{Radial polynomial envelope degree $p$} & $[6.0, 6.0, 6.0]$ \\
\midrule[0.6pt]
Maximum degree $L_{\text{max}}$ & $3$ \\
Number of convolutional layers & $5$ \\
Hidden Irreps & $128\times0e + 64\times1o + 32\times2e + 32\times3o$ \\
Number of invariant layers & $2$ \\
Number of invariant neurons & $128$ \\
Stress gradient mode & Edge-based \\
\bottomrule[1.2pt]
\end{tabular}
}
\vspace{1mm}
\label{appendix:tab:eqnorm_hyperparameters}
\end{table}

All models adopt independent radial MLPs and envelope functions for each cutoff, with learnable scale-mixing coefficients $\gamma_n$. Implementation was built upon the official codebases with necessary extensions for multi-cutoff support.

\clearpage 
\newpage
\section*{NeurIPS Paper Checklist}

The checklist is designed to encourage best practices for responsible machine learning research, addressing issues of reproducibility, transparency, research ethics, and societal impact. Do not remove the checklist: {\bf The papers not including the checklist will be desk rejected.} The checklist should follow the references and follow the (optional) supplemental material.  The checklist does NOT count towards the page
limit. 

Please read the checklist guidelines carefully for information on how to answer these questions. For each question in the checklist:
\begin{itemize}
    \item You should answer \answerYes{}, \answerNo{}, or \answerNA{}.
    \item \answerNA{} means either that the question is Not Applicable for that particular paper or the relevant information is Not Available.
    \item Please provide a short (1--2 sentence) justification right after your answer (even for \answerNA). 
\end{itemize}

{\bf The checklist answers are an integral part of your paper submission.} They are visible to the reviewers, area chairs, senior area chairs, and ethics reviewers. You will also be asked to include it (after eventual revisions) with the final version of your paper, and its final version will be published with the paper.

The reviewers of your paper will be asked to use the checklist as one of the factors in their evaluation. While \answerYes{} is generally preferable to \answerNo{}, it is perfectly acceptable to answer \answerNo{} provided a proper justification is given (e.g., error bars are not reported because it would be too computationally expensive'' or ``we were unable to find the license for the dataset we used''). In general, answering \answerNo{} or \answerNA{} is not grounds for rejection. While the questions are phrased in a binary way, we acknowledge that the true answer is often more nuanced, so please just use your best judgment and write a justification to elaborate. All supporting evidence can appear either in the main paper or the supplemental material, provided in appendix. If you answer \answerYes{} to a question, in the justification please point to the section(s) where related material for the question can be found.

IMPORTANT, please:
\begin{itemize}
    \item {\bf Delete this instruction block, but keep the section heading ``NeurIPS Paper Checklist"},
    \item  {\bf Keep the checklist subsection headings, questions/answers and guidelines below.}
    \item {\bf Do not modify the questions and only use the provided macros for your answers}.
\end{itemize}


\begin{enumerate}

\item {\bf Claims}
    \item[] Question: Do the main claims made in the abstract and introduction accurately reflect the paper's contributions and scope?
    \item[] Answer: \answerYes{} 
    \item[] Justification: We assure that the main claims made in the abstract and introduction accurately
reflect the paper’s contributions and scope.

\item {\bf Limitations}
    \item[] Question: Does the paper discuss the limitations of the work performed by the authors?
    \item[] Answer: \answerYes{} 
    \item[] Justification: The limitations are in  Appendix \ref{appx:Limitations}.

\item {\bf Theory assumptions and proofs}
    \item[] Question: For each theoretical result, does the paper provide the full set of assumptions and a complete (and correct) proof?
    \item[] Answer: \answerYes{} 
    \item[] Justification: Proofs and Derivations are in  Appendix.

    \item {\bf Experimental result reproducibility}
    \item[] Question: Does the paper fully disclose all the information needed to reproduce the main experimental results of the paper to the extent that it affects the main claims and/or conclusions of the paper (regardless of whether the code and data are provided or not)?
    \item[] Answer: \answerYes{} 
    \item[] Justification: We provide details for reproduction in Section \ref{sec:experiments} and Appendix \ref{appx:details}.

\item {\bf Open access to data and code}
    \item[] Question: Does the paper provide open access to the data and code, with sufficient instructions to faithfully reproduce the main experimental results, as described in supplemental material?
    \item[] Answer: \answerYes{} 
    \item[] Justification: The code is available at the anonymous repository:  {\urlstyle{same}\hypersetup{allcolors=magenta}\url{https://anonymous.4open.science/r/MCSD-C854}}.
    

\item {\bf Experimental setting/details}
    \item[] Question: Does the paper specify all the training and test details (e.g., data splits, hyperparameters, how they were chosen, type of optimizer) necessary to understand the results?
    \item[] Answer: \answerYes{} 
    \item[] Justification: The details are provided in Section \ref{sec:experiments} and Appendix \ref{appx:details}.

\item {\bf Experiment statistical significance}
    \item[] Question: Does the paper report error bars suitably and correctly defined or other appropriate information about the statistical significance of the experiments?
    \item[] Answer: \answerNo{} 
    \item[] Justification: Our experiments are mostly computational expensive. Therefore, we run only
one group of experiment for each setting, following previous works.

\item {\bf Experiments compute resources}
    \item[] Question: For each experiment, does the paper provide sufficient information on the computer resources (type of compute workers, memory, time of execution) needed to reproduce the experiments?
    \item[] Answer: \answerYes{} 
    \item[] Justification: The details are provided in Section \ref{sec:experiments} and Appendix \ref{appx:details}.
    
\item {\bf Code of ethics}
    \item[] Question: Does the research conducted in the paper conform, in every respect, with the NeurIPS Code of Ethics \url{https://neurips.cc/public/EthicsGuidelines}?
    \item[] Answer: \answerYes{} 
    \item[] Justification: We respect the NeurIPS code of ethics.

\item {\bf Broader impacts}
    \item[] Question: Does the paper discuss both potential positive societal impacts and negative societal impacts of the work performed?
    \item[] Answer: \answerNA{} 
    \item[] Justification: This work is about physics/materials science, which does not involve
societal impact.
    
\item {\bf Safeguards}
    \item[] Question: Does the paper describe safeguards that have been put in place for responsible release of data or models that have a high risk for misuse (e.g., pre-trained language models, image generators, or scraped datasets)?
    \item[] Answer: \answerNA{} 
    \item[] Justification: Our research is about physics/materials science that dose not have such
risks.

\item {\bf Licenses for existing assets}
    \item[] Question: Are the creators or original owners of assets (e.g., code, data, models), used in the paper, properly credited and are the license and terms of use explicitly mentioned and properly respected?
    \item[] Answer: \answerYes{} 
    \item[] Justification: The discussion is in Appendix \ref{appx:details}.

\item {\bf New assets}
    \item[] Question: Are new assets introduced in the paper well documented and is the documentation provided alongside the assets?
    \item[] Answer: \answerYes{} 
    \item[] Justification: The code is available at the anonymous repository:  {\urlstyle{same}\hypersetup{allcolors=magenta}\url{https://anonymous.4open.science/r/MCSD-C854}} with a README as the documentation.

\item {\bf Crowdsourcing and research with human subjects}
    \item[] Question: For crowdsourcing experiments and research with human subjects, does the paper include the full text of instructions given to participants and screenshots, if applicable, as well as details about compensation (if any)? 
    \item[] Answer: \answerNA{} 
    \item[] Justification: We do not include these resources.

\item {\bf Institutional review board (IRB) approvals or equivalent for research with human subjects}
    \item[] Question: Does the paper describe potential risks incurred by study participants, whether such risks were disclosed to the subjects, and whether Institutional Review Board (IRB) approvals (or an equivalent approval/review based on the requirements of your country or institution) were obtained?
    \item[] Answer: \answerNA{} 
    \item[] Justification: We do not include such risks.

\item {\bf Declaration of LLM usage}
    \item[] Question: Does the paper describe the usage of LLMs if it is an important, original, or non-standard component of the core methods in this research? Note that if the LLM is used only for writing, editing, or formatting purposes and does \emph{not} impact the core methodology, scientific rigor, or originality of the research, declaration is not required.
    \item[] Answer: \answerNA{} 
    \item[] Justification: The core method development in this research does not involve LLMs as any
important, original, or non-standard components.

\end{enumerate}

\end{document}